\documentclass{article} 
\usepackage{iclr2015,times}
\usepackage{hyperref}
\usepackage{url}
\usepackage{amsmath}
\usepackage{amsfonts}
\usepackage{multirow}
\usepackage{url}
\usepackage{graphicx} 

\usepackage{boxedminipage}  
\usepackage{color}
\usepackage{colortbl}
\usepackage{enumitem}

\newcommand{\mention}[1]{\textit{#1}}
\newcommand{\entity}[1]{\texttt{#1}}

\def\bs#1{\boldsymbol{#1}}

\title{Inducing Semantic Representation from Text by Jointly Predicting and
Factorizing Relations}

\author{
Ivan Titov, Ehsan Khoddam \\
Universiteit van Amsterdam\\
Amsterdam, the Netherlands \\
\texttt{\{titov,e.khoddammohammadi\}@uva.nl} \\
}

\iclrfinalcopy 

\begin{document}

\maketitle

\begin{abstract}

In this work, we propose a new method to integrate two recent lines of work:
unsupervised induction of shallow semantics (e.g., semantic roles) and factorization of relations in text and knowledge bases.  
Our model consists of two components: (1) an encoding component: a semantic role labeling
model which predicts roles given a rich set of syntactic and lexical features; (2) a reconstruction component:
a tensor factorization model which relies on roles to predict argument fillers. 
When the components are estimated jointly to minimize
errors in argument reconstruction,  the induced roles largely correspond to roles defined in annotated resources.
Our method performs on par with most accurate  role induction methods on English, even though, unlike these previous approaches,  we do not
incorporate any prior linguistic knowledge about the language.
\end{abstract}

\section{Introduction}

Shallow representations of  meaning,  and  semantic  role  labels  in
particular,  have a  long history  in  linguistics \citep{fillmore-68}.
More recently, with an emergence of large annotated resources such as
PropBank \citep{Palmer05} and FrameNet \citep{Baker98}, automatic semantic role
labeling (SRL) has  attracted a lot of
attention~\citep{conll08,conll09,Das2010}.

Semantic role representations encode the underlying predicate-argument structure of sentences, or,
more specifically, for every predicate in a sentence they identify
a set of arguments and associate each argument with an underlying {\it semantic
role}, such as an agent (an initiator or doer of the action) or a  patient
(an affected entity).
Consider the following sentence:
\begin{description}
\item[] $[_{Agent}$ The police$]$ charged $[_{Patient}$ the demonstrators$]$  $[_{Instrument}$ with batons$]$.
\end{description}
Here, {\it the police},  {\it the demonstrators} and {\it with batons} are assigned to roles {\entity Agent}, {\entity Patient} and {\entity Instrument}, respectively.
Semantic roles have many potential applications in NLP and have  been shown to
benefit, for example, question answering \citep{Shen07,berant2014modeling} and
textual entailment \citep{Sammons09}, among others.

The scarcity of 
annotated data has motivated the research into unsupervised learning of semantic representations~\citep{Lang10,Lang11a,Lang11b,TitovKlementEacl12,furstenau2012,garg2012}. 
The existing methods have a number of serious shortcomings.  First, they make very strong assumptions, for example, assuming
that arguments  are conditionally independent of each other given the predicate. Second, unlike state-of-the-art supervised parsers, they rely on a very simplistic set of features of a sentence. 
These factors lead to models being insufficiently expressive to capture syntax-semantics interface,  inadequate handling of  language ambiguity and, overall, introduces an upper bound on their performance.
 



 In this work, we propose a method  for  effective unsupervised estimation of feature-rich models of semantic roles.  We demonstrate that reconstruction-error objectives, which have been shown to be effective primarily for training neural networks, 
are well suited for inducing feature-rich log-linear models of semantics. Our model consists of two components: a log-linear feature rich semantic role labeler and
a tensor-factorization model which captures interaction between semantic roles and argument fillers. 
Our method rivals the most accurate 
semantic role induction methods on English~\citep{TitovKlementEacl12,Lang11a}. 
Importantly, no prior knowledge about any specific language was incorporated in
our feature-rich model, whereas the clustering counterparts relied on language-specific argument signatures.




\section{Approach}


At the core of our approach is a 
statistical model encoding an interdependence between a semantic role structure
and its realization in a sentence. In the unsupervised learning setting, 
sentences, their syntactic representations and argument positions 
(denoted by $x$) are observable whereas the associated semantic roles $\bs{r}$
are latent and need to be induced by the model. Crucially, the good $\bs{r}$
should encode roles rather than some other form of abstraction.
In what follows, we will refer to roles using their names, though, in the unsupervised setting, our method, as any other latent variable model, will not yield human-interpretable labels for them.  
We also focus only on the labeling stage
of semantic role labeling.
Identification, though an important problem, can be
tackled   with heuristics~\citep{Lang11a}, with 
unsupervised techniques~\citep{Abend09} or potentially by using a supervised classifier
trained on a small amount of data.

The model consists of two components.
The first component is responsible for prediction of argument tuples based on roles and the predicate. 
In our experiments,  in this component, we  represent arguments as lemmas of their lexical heads (e.g., {\it baton} instead of {\it with batons}), and we also
restrict ourselves to only verbal predicates. 
Intuitively, we can think of predicting one argument at a time (see Figure~1(b)): an argument
 (e.g., \mention{demonstrator} in our example) is predicted based on the predicate lemma (\mention{charge}), 
the role assigned to this argument (i.e. \entity{Patient}) and other role-argument pairs ((\entity{Agent}, \mention{police}) and
(\entity{Instrument}, \mention{baton})).  While learning to predict arguments, the inference algorithm will search for role assignments which 
simplify this prediction task as much as possible. Our hypothesis is that these assignments will correspond to 
roles 
 accepted in linguistic theories (or, more importantly, useful in practical applications). 
Why is this hypothesis plausible? 
Primarily because these semantic representations were introduced as an abstraction capturing crucial properties of a relation (or an event). Thus, these representations, rather than surface linguistic details like argument order or syntactic functions, should be crucial for modeling sets of potential argument tuples.
%
%
The reconstruction component is not the only  part of the model.
Crucially, what we referred to above as 
`searching for role assignments to simplify argument prediction'  would actually correspond to learning another component: a 
semantic role labeler which predicts roles relying on a rich set of sentence features. These two components will be estimated jointly in such a way as to minimize errors in recovering arguments.  The role labeler will be the end-product of learning: it will be used to process new sentences, and it will be compared to existing methods in our evaluation.

\begin{figure*}
\label{fig:reconstr}
\begin{center}
\includegraphics[width=\columnwidth]{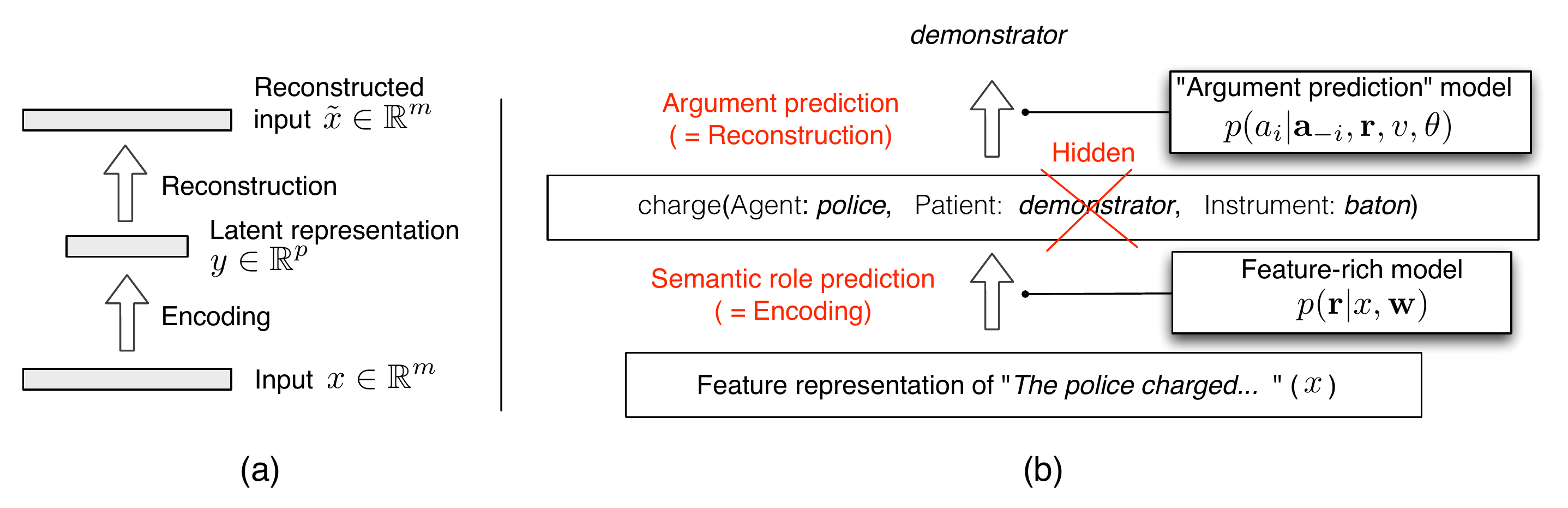}
\caption{(a) An autoencoder from $\mathbb{R}^m$ to $\mathbb{R}^p$ (typically $p < m$).  ~~ (b) Modeling roles within the reconstruction-error minimization framework.
} 
\end{center}
\end{figure*}

Generative modeling is not the only way  to learn latent representations.
One alternative, popular in the neural network community, is to use autoencoders
instead and optimize the reconstruction error~\citep{hinton1989,vincent2008}.
The encoding model will be a feature-rich classifier which predicts semantic roles for a sentence, and the reconstruction model is the model which predicts an argument given its role, and given the rest of the arguments and their roles. 
The idea of training linear models with reconstruction error was previously explored by~\citet{daume2009} and very recently by~\citet{ammar2014}. 
However, they do not consider learning factorization models, and they also do
not deal with semantics.    
Tensor and factorization methods used in the context of  modeling knoweldge bases
(e.g., \citep{bordes2011}) are also close in spirit. However, they do not deal with inducing semantics but rather factorize existing relations (i.e. rely on semantics).

\subsection{Modeling semantics within the reconstruction-error framework}
As we mentioned above, we focus on argument
labeling: we assume that arguments $\bs{a}  =
(a_1, \ldots, a_N)$, $a_i \in \mathcal{A}$,
 are known, and only their roles $\bs{r}  = (r_1, \ldots, r_N)$, $r_i \in \mathcal{R}$ need to be induced. 
For the encoder (i.e. the semantic role labeler), we use a log-linear model:
\begin{equation}
\nonumber
p(\bs{r}| x, \bs{w}) \propto  
\exp(\bs{w}^T \bs{g}(x, \bs{r})),
\end{equation}
where $\bs{g}(x, \bs{r})$ is a feature vector 
encoding interactions between sentence $x$ and the semantic
role representation $\bs{r}$. Any model can be used here as long as
the posterior distributions of roles $r_i$  can
be efficiently computed or approximated. 
In our experiments, we used a model which
factorizes over individual arguments (i.e. independent logistic regression classifiers).
  
The reconstruction component predicts an argument (e.g.,  the $i$th argument
$a_i$) given the semantic roles $\bs{r}$,  the predicate  $v$ and other
arguments $\bs{a}_{-i} =  (a_1, \ldots, a_{i-1}, a_{i+1}, \ldots, a_N)$ with a bilinear softmax model:
\begin{equation}
\label{expr:reconstr}
p(a_i | \bs{a}_{-i},\bs{r}, v, C, \bs{u}) = \frac{\exp({  \bs{u}_{a_i}^T 
C_{v,r_i}^T   \sum_{j \neq i} { C_{v,r_j} \bs{u}_{a_{j}} }})}{ Z(\bs{r}, v, i)},
\end{equation}
 $\bs{u}_{a} \in \mathbb{R}^d$ (for every $a \in \mathcal{A}$)  and $C_{v,r} \in \mathbb{R}^{d \times k}$ (for every verb $v$ and every role $r \in \mathcal{R}$)
 are model parameters, $Z(\bs{r}, v, i)$ is the partition function ensuring that the probabilities sum to one.
Intuitively, embeddings $\bs{u}_a$ 
 encode semantic properties of an argument: for example, embeddings for the words \mention{demonstrator} and \mention{protestor} should be somewhere near each other in $\mathbb{R}^d$ space, and further away from that for the word \mention{cat}. The product $C_{p,r} \bs{u}_a$ is a $k$-dimensional vector encoding beliefs about other arguments based on the argument-role pair 
$(a, r)$. 
In turn, the dot product $(C_{v,r_i} \bs{u}_{a_i})^T C_{v,r_j} \bs{u}_{a_j}$ is
large if the argument pair $(a_i, a_j)$ is semantically compatible with the
predicate, and small otherwise.
Intuitively, this objective corresponds to scoring argument tuples according to
\begin{equation}
\vspace{-1ex}
\nonumber
h(\bs{a}, \bs{r}, v, C, \bs{u}) = \sum_{i \neq j} {\bs{u}_{a_i}^T C^T_{v,r_i} C_{v,r_j} \bs{u}_{a_j}},
\end{equation}
hinting at connections to
(coupled) tensor and factorization methods~\citep{yilmaz2011,bordes2011} and distributional semantics~ \citep{mikolov2013efficient,pennington2014}.
Note also that the reconstruction model does not have 
access to any features of the sentence
 (e.g., argument order or syntax), forcing the roles to convey all the necessary information. 
 
 
 
 
In practice, we smooth the model
by using a sum of predicate-specific and cross-predicate projection matrices $(C_{v,r} + C_{r})$ instead of just $C_{v,r}$.


\subsection{Learning}
\label{sect:learning}

Parameters of both model components ($\bs{w}$, $\bs{u}$ and $C$)  are  learned jointly: the 
natural objective associated with every sentence would be the following:
\begin{equation}
\label{expr:unlabobj}
 \sum_{i=1}^N {\log \sum_{\bs{r}} { p(a_i | \bs{a}_{-i}, \bs{r}, v, C, \bs{u}) p(\bs{r} | x, \bs{w})}}.
\end{equation}
However optimizing this objective is not practical in its exact form for two 
reasons: (1) the marginalization over $\bs{r}$ is exponential in the number of
arguments; (2) the partition function $Z(\bs{r}, v, i)$ requires summation over the entire set of potential argument lemmas. We use existing techniques
to address both challenges.   
In order to deal with the first challenge, 
we use a basic mean-field approximation: instead of marginalization over $r$ we
substitute $r$ with their posterior distributions $\mu_{is} = p(\bs{r_i} = s | x, \bs{w})$.
%
%
%
To tackle the second problem,  the computation of 
$Z(\bs{\mu}, v, i)$, we use a negative sampling technique (see, e.g.,~\citet{mikolov2013efficient}).
At test time, only the linear semantic role labeler is used, so the inference is straightforward.

%
\vspace{-1ex}
\section{Experiments}
\label{sec:exp}
 \vspace{-1ex}

We followed~\citet{Lang10} and used the CoNLL 2008 shared task data~\citep{conll08}. 
%
As in most previous work on unsupervised SRL, we evaluate our model using clustering metrics: purity, collocation and their harmonic mean F1. 
%
%
For the semantic role labeling (encoding) component,
we relied on 14 feature patterns used for argument labeling in one of the
popular supervised role labelers~\citep{johansson2008}, which resulted in a
quite large feature space (49,474 feature instantiations for our English dataset).

For the reconstruction component, we set the dimensionality of embeddings $d$,  
the projection dimensionality $k$ and the number of negative samples $n$ to
$30$, $15$ and $20$, respectively. These hyperparameters were tuned on held-out
data (the CoNLL 2008 development set), $d$ and $k$ were chosen among $\lbrace
10,15,20,30,50\rbrace$ with constraining $d$ to be always greater than $k$, $n$
was fixed to $10$ and $20$.
The model was not sensitive to the parameter defining the number of roles as long it was large enough. 
For training, we used uniform random initialization and AdaGrad~\citep{duchi2011adaptive}.

 
Following ~\citep{Lang10}, we use a baseline ({\em SyntF}) which simply
clusters predicate arguments according to the dependency relation to their head.
A separate cluster is allocated for each of 20 
most frequent relations in the dataset and an additional cluster is used for all
other relations.
As observed in the previous work~\citep{Lang11a}, this is a hard baseline to beat. 

We also compare against previous approaches:
the latent logistic classification model \citep{Lang10}  (labeled {\em LLogistic}), 
the  agglomerative clustering  method \citep{Lang11a} ({\em Agglom}),  the graph partitioning approach \citep{Lang11b} ({\em GraphPart}),
the global role ordering model~\citep{garg2012} ({\em RoleOrdering}). We also report results of an  improved version of {\em Agglom},
recently reported by~\citet{Lang14} ({\em Agglom+}).
The strongest previous model is {\em Bayes}: {\em Bayes} is the most accurate (`coupled') version of  
the Bayesian model of~\citet{TitovKlementEacl12},
estimated from the CoNLL data without relying on any external data.
 
Our model outperforms or performs on par with best previous models in terms of F1 (see Table~1). Interestingly, the purity and collocation balance is very different for our model and for the rest of the systems.
In fact, our model induces at most 4-6 roles.  On the contrary,  {\em Bayes} predicts more than 30 roles for the majority of frequent predicates (e.g., 43 roles for the predicate {\it include} or 35 for {\it say}).
Though this tendency reduces the purity scores for our model, this also means that our roles are more human interpretable. For example, agents and patients are clearly identifiable in the model predictions.  


\begin{table}[t]
\caption{Purity (PU), collocation (CO) and F1 on English (PropBank / CoNLL
2008).}
\begin{center}

\begin{tabular} {l c c c}  

\label{tab:en}
                & PU & CO & F1 \\
\hline
Our Model         & 79.7 & 86.2 & {\bf 82.8}   \\
Bayes           & 89.3 & 76.6 & 82.5   \\
Agglom+          &  87.9 & 75.6 & 81.3 \\
RoleOrdering    & 83.5 & 78.5 & 80.9 \\
Agglom        & 88.7 & 73.0 & 80.1   \\
GraphPart		  &	88.6 & 70.7 & 78.6   \\
LLogistic         & 79.5 & 76.5 & 78.0   \\

\hline
SyntF		      & 81.6 & 77.5 & 79.5   \\
\end{tabular}

\end{center}
\end{table}

\vspace{-1ex}
\section{Conclusions}
\vspace{-1ex}

We introduced a method for inducing feature-rich semantic role labelers from
unannoated text.
 In our approach, we view 
a semantic role representation as an encoding of a latent relation between a
predicate and a tuple of its arguments. We capture this relation with a probabilistic tensor factorization model.  
Our estimation method yields a semantic role labeler which achieves
state-of-the-art results on English.

{\footnotesize \bibliography{iclr2015}}
\bibliographystyle{apalike}
\end{document}